\documentclass[letterpaper, 10 pt, conference]{ieeeconf}  

\IEEEoverridecommandlockouts                              

\usepackage[T1]{fontenc}

\usepackage{amsfonts}       
\usepackage{nicefrac}       
\usepackage{subfig}			
\usepackage{booktabs}	
\usepackage[linesnumbered,ruled,vlined]{algorithm2e}
\usepackage{xcolor}
\usepackage[normalem]{ulem} 
\usepackage{mathtools}
\usepackage[percent]{overpic} 
\usepackage{comment}
\usepackage[noadjust]{cite}

\definecolor{black}{rgb}{0, 0, 0}
\definecolor{red}{rgb}{0.9, 0, 0}
\definecolor{green}{rgb}{0, 0.6, 0}
\definecolor{blue}{rgb}{0, 0, 0.9}
\definecolor{grey}{rgb}{0.52, 0.52, 0.51}

\newcommand{\RED}[1]{\textcolor{red}{#1}}

\newcommand{\rev}[1]{\textcolor{black}{#1}}

\newcommand{\mde}[0]{\texttt{MDE}}
\newcommand{\mabrrt}[0]{\texttt{MAB-RRT}}
\newcommand{\ctxrrt}[0]{\texttt{CTX-RRT}}
\newcommand{\rrt}[0]{\texttt{RRT}}
\newcommand{\MABRRT}[0]{\mabrrt}
\newcommand{\RRT}[0]{\rrt}
\newcommand{\CTXRRT}[0]{\ctxrrt}

\SetKwInOut{KwParameter}{Parameters}
\SetKw{KwContinue}{continue}
\SetKw{KwBreak}{break}
\SetKw{KwInterval}{in}
\SetKw{KwOr}{or}
\SetKw{KwAnd}{and}
\SetKwFunction{unif}{unif}

\newcommand{\X}[0]{X}
\newcommand{\x}[0]{x}
\newcommand{\U}[0]{U}

\newcommand{\fillbox}[3]
{\bgroup
  \dimen1=#1\relax
  \dimen2=#2\relax
  \sbox0{\includegraphics[width=#1]{#3}}%
  \ifdim\ht0>\dimen2
    \dimen0=\dimexpr \ht0-\dimen2\relax
    \adjustbox{clip=true,trim=0pt 0.5\dimen0 0pt 0.5\dimen0}{\usebox0}%
  \else
    \sbox0{\includegraphics[height=#2]{#3}}%
    \ifdim\wd0>\dimen1
      \dimen0=\dimexpr \wd0-\dimen1\relax
      \adjustbox{clip=true,trim=0.5\dimen0 0pt 0.5\dimen0 0pt}{\usebox0}%
    \else
      \usebox0
    \fi
  \fi
\egroup}

\newtheorem{problem}{Problem}

\title{\LARGE \bf
Online Adaptation of Sampling-Based Motion Planning with\\ Inaccurate Models
}
\author{Marco Faroni and Dmitry Berenson 
\thanks{This work was supported in part by the Office of Naval Research under Grant N00014-21-1-2118, and in part by the NSF under Grants IIS-1750489, IIS-2113401, and IIS-2220876.
The authors are with the Robotics Department, University of Michigan, Ann Arbor, MI 48109, United States. {\tt\footnotesize \{mfaroni; dmitryb\}@umich.edu}}
}

\begin{document}

\maketitle
\thispagestyle{empty}
\pagestyle{empty}

\begin{abstract}
Robotic manipulation relies on analytical or learned models to simulate the system dynamics.
These models are often inaccurate and based on offline information, so that the robot planner is unable to cope with mismatches between the expected and the actual behavior of the system (e.g., the presence of an unexpected obstacle).
In these situations, the robot should use information gathered online to correct its planning strategy and adapt to the actual system response.
We propose a sampling-based motion planning approach that uses an estimate of the model error and online observations to correct the planning strategy at each new replanning.
Our approach adapts the cost function and the sampling bias of a kinodynamic motion planner when the outcome of the executed transitions is different from the expected one (e.g., when the robot unexpectedly collides with an obstacle) so that future trajectories will avoid unreliable motions.
To infer the properties of a new transition, we introduce the notion of context-awareness, i.e., we store local environment information for each executed transition and avoid new transitions with context similar to previous unreliable ones.
This is helpful for leveraging online information even if the simulated transitions are far (in the state-and-action space) from the executed ones.
Simulation and experimental results show that the proposed approach increases the success rate in execution and reduces the number of replannings needed to reach the goal.
\end{abstract}

\section{Introduction}

\begin{figure}[tpb]
	\centering
	\includegraphics[trim = 1.0cm 6cm 2.2cm 1.0cm, clip, angle=0, width=0.8\columnwidth, height=7cm]{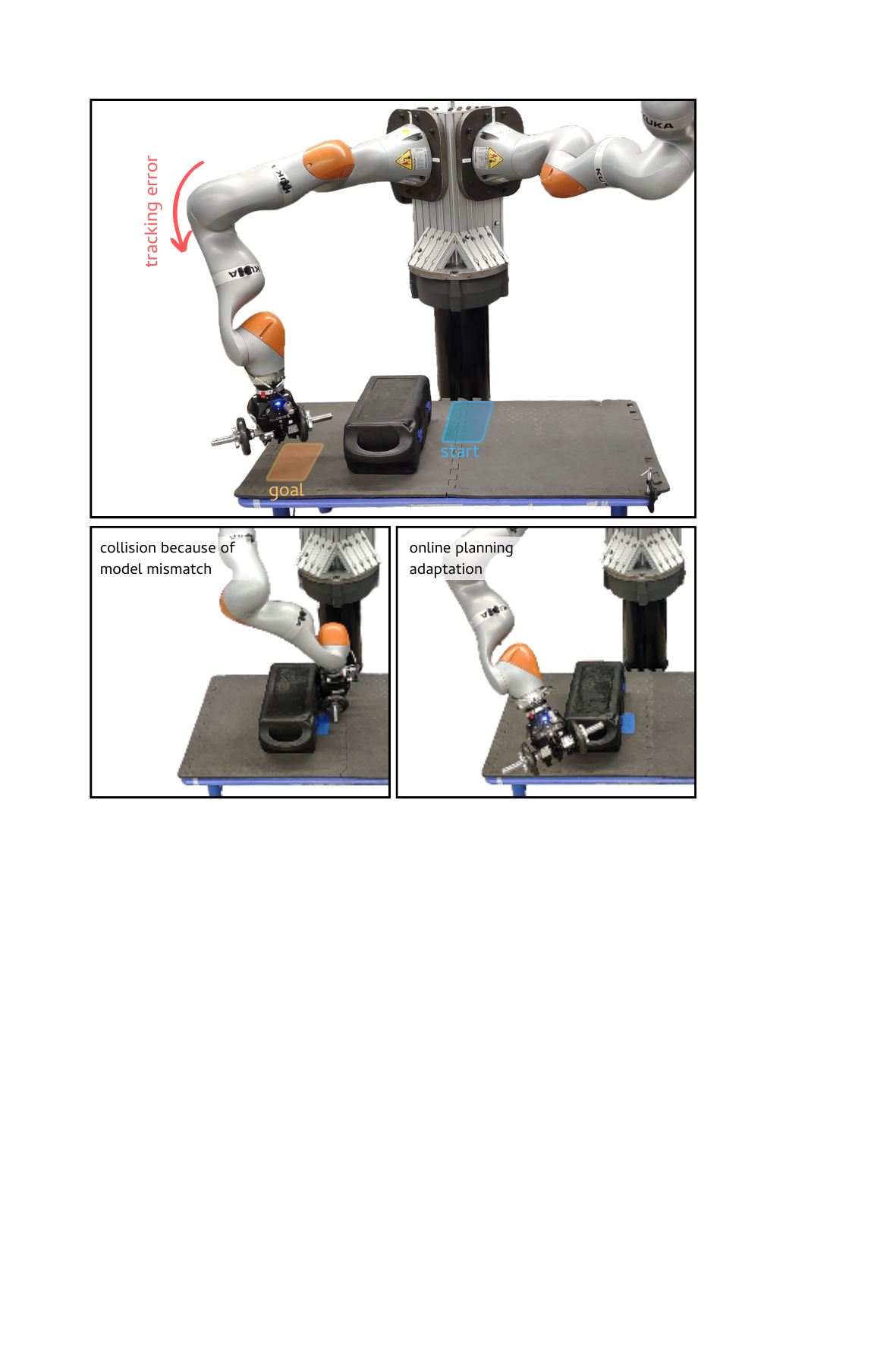}
	\caption{A 7-degree-of-freedom manipulator carrying a weight with uncertain tracking control.}
	\label{fig: victor-dumbbell}
 \vspace{-0.7cm}
\end{figure}

The advances in physics-based simulation and deep-learning models in robotics allow to achieve complex tasks such as manipulation of deformable objects and liquids \cite{Lippi:visual-action-planning,NICOLA2024102630,Mitrano:science-robotics,corl-planning-with-spatial-temporal-abstraction,Mitrano:focus-adaptation,fluid-manipulation,coffee-steering}, contact-rich manipulation \cite{Kromere:contact-rich-manipulation,brock-contact-based-rrt,motion-planning-sliding} and safe, compliant manipulation \cite{garabini:motion-planning-soft-robot,marcucci2020parametric}.
These models are often inaccurate in predicting the outcome of actions (e.g., because of the epistemic uncertainty of learned models or the sim-to-real gap of physics simulators).
For example, consider the tabletop scenario in Fig. \ref{fig: victor-dumbbell}: A compliant manipulator moves a heavy object across the table; because of the payload and the compliant control, the trajectory execution will deviate from the planned path, possibly causing unexpected collisions.
Previous works addressed this problem by propagating uncertainty throughout the search \cite{planning-belief-space,planning-uncertainty-dynamic-programming} or by discarding actions deemed to lead to significant errors \cite{Mitrano:science-robotics}.
This usually requires complex modeling of the system uncertainty over the whole state-and-action space.
However, even minor mismatches between the offline and the online scenario can jeopardize the effectiveness of the uncertainty-aware planner.
In the example of Fig. \ref{fig: victor-dumbbell}, a mismatch between the simulated and real compliance would cause the planner to find trajectories that are infeasible in the real world.
Suppose the real dumbbell is heavier than the simulated one; then, the trajectories would often cause the robot to collide with the environment.
Even in the case of replanning, the planner would likely continue generating a trajectory that causes a collision if it does not take online data into account. 
Updating the system model according to the observed behavior can be computationally demanding (e.g., it could require retraining a deep network representing the dynamics) and often cannot be performed online. 
Adaptive methods are typically restricted to short-horizon planning and locally update the system dynamics in the neighborhood of the current configuration when the robot is stuck \cite{CMAX,CMAX++,Johnson:tampc}.
As a shortcoming, these methods adapt the structure of the planning problem only in a small neighborhood of the executed transitions.
For example, \cite{CMAX} and \cite{CMAX++} penalize state-and-action pairs within a certain distance from previous inaccurate transitions.
That is, the planner will attempt to avoid transitions in a small neighborhood of the current and previous robot configurations.

This work aims to bridge the gap between long-horizon sampling-based planning and online model adaptation to provide a robust and adaptive solution to motion planning with inaccurate models.
The main idea is to formulate the problem as an optimal motion planning problem, minimizing the expected deviation between the robot model and the real system. 
Such expected deviation includes an offline estimate of the model inaccuracy and a residual term obtained from previously executed trajectories.
In particular, the residual term discourages transitions similar to those that failed in previous executions.
Similarly, we adapt the sampling bias of the planner to discourage sampling such transitions.
To do so, we introduce the notion of  context-aware similarity, which allows the planner to infer whether new simulated transitions will be unreliable during the execution.
In our examples, the context of a transition is composed of the offline-estimated model deviation and the presence of obstacles in the neighborhood of the transition.
Under the assumption that transitions with a similar context will yield a similar outcome (even if they are far in the state-and-action space), the planner can infer the reliability of new simulated transitions and adapt the motion planning problem after a few executions.
To do so, we update the cost function by assigning a penalty term according to the probability that a transition is unreliable (i.e., if it has a context similar to previous unreliable transitions).
Then, we update the sampling distribution of our RRT-based planner to undersample transitions with high probability of being unreliable.
To adapt the sampling distribution online, we build on the adaptive sampling proposed in \cite{Faroni_RAL2023}, which clusters previous transitions and uses a Multi-Armed Bandit-based sampler to draw new transitions from such clusters.
\cite{Faroni_RAL2023} clusters simulated transitions with respect to their cost and state-space distance.
In this work, we cluster them with respect to their local context and penalize clusters containing unreliable transitions in the Multi-Armed Bandit (MAB).
The result is that the planner discourages possible unreliable transitions both at sampling time and when it evaluates the cost function.
We demonstrate that the proposed approach increases the execution success rate and reduces the number of necessary replannings to reach the goal in 2D examples and a 7-degree-of-freedom manipulation scenario.


\section{Related Works}\label{sec:related-works}

Motion planning under uncertainty is a long-studied topic in robotics.
The goal is to find a sequence of motions that drive the robot from a start to a goal configuration despite a flawed representation of the robot and/or the environment.
Because sampling-based planners such as RRT \cite{lavalle:rrt-ijrr} and PRM \cite{Kavraki:PRM} are the most widespread in robotics, a wide variety of sampling-based planners have been proposed also to cope with uncertainty.
The most common approach is to consider an estimate of the model uncertainty while building the search tree. 
For example, the uncertainty estimate can be used to plan in a belief space that models the distribution of the possible outcomes of a robot action \cite{planning-belief-space,planning-belief-2022,Pavone:robust-rrt}.
The shortcoming is that such planners fail when it is impossible to find a solution under all possible realizations of the uncertainty model.
Moreover, they often approximate the set of reachable states in a sampling-based fashion, requiring a large number of dynamics evaluations at each iteration.

Mitrano et al. \cite{Mitrano:science-robotics} proposed a different approach that learns an estimate of the model deviation to avoid solutions with large expected error.
It learns a classifier that predicts whether a transition will be reliable or not when executed and uses it to discard unreliable transitions in an RRT.
Similarly, \cite{Faroni_RAL2023} uses an estimate of the model error as a cost function in an RRT-like planner and shows that low-cost solutions lead to a higher success rate.
Other approaches aim to find a trusted domain where a given controller can compensate for the execution error \cite{Glen:trusted-domain}.
Replanning approaches have also proved effective in coping with uncertain dynamics \cite{TRO-fast-replanning} and moving obstacles \cite{Faroni_IEEE_ACCESS}.
All these approaches rely on offline information to quantify the model uncertainty. However, they do not reason about possible mismatches between the uncertainty estimate and the real-world outcome of the robot actions.
\rev{That is, even if a robust planner is used, these methods do not leverage new observations to correct the behavior of the planner, possibly getting the robot stuck repeatedly.}
\rev{Moreover, updating the system model according to the observed behavior is prohibitive because it requires retraining the model after each new observation. 
}

Outside the realm of sampling-based planning, other approaches addressed the possibility of changing the planning strategy online according to online information on the robot execution error.
For example, \cite{CMAX} and \cite{CMAX++} use A$^*$ to locally penalize the neighborhood of state-and-action pairs that proved to be unreliable (i.e., such that their actual execution error was different from the expected one). 
This prevents repeating unreliable actions at the next execution.
However, the correction only applies to states and actions in a small neighborhood of the executed ones and A$^*$ is usually inefficient with large continuous spaces, as is typical of robotic manipulation. 
As for online planning, receding-horizon planners have been proposed to deal with errors in execution. 
E.g., \cite{Johnson:tampc} implements a receding-horizon planner that trades off the avoidance of states with unexpected dynamics and stagnation of the progress towards the goal.

This work focuses on sampling-based kinodynamic motion planning.
In particular, we would like a kinodynamic planner to adapt the search according to an estimate of the model error and online observations gathered during the execution.
To do so, we update the cost function and the sampling distribution of the planner.
Biasing the sampling is a common strategy to improve the solution cost quickly \cite{gammell:review,amato:obprm,amato:evaluating-guiding-spaces,Pavone:learning-sampling,Burdick:learning-sampling-distribution,Kingston:learning-sampling,Laumond:pca-rrt}. 
Sampling bias may be learned \textit{a priori} \cite{Pavone:learning-sampling, Burdick:learning-sampling-distribution} or adapted during the planning \cite{Laumond:pca-rrt,Faroni_RAL2023}.
We leverage the online-learning approach described in \cite{Faroni_RAL2023} to trade off exploitation of regions of the search space with low cost vs. less-explored ones.
\cite{Faroni_RAL2023}  runs \RRT\ multiple times and, at the end of each run, it groups the transitions via clustering.
Then, it uses a Multi-Armed Bandit (MAB) algorithm to decide whether the next transition will be drawn from a cluster or the uniform distribution.
We leverage this sampling strategies with two major changes: (i) we perform clustering with respect to the context of the transitions instead of using state-and-action space distance; (ii) we modify the MAB's rewards of the clusters that contain unreliable transitions.

\section{Problem Statement}\label{sec:problem-statement}

Consider a dynamical system, $x_{k+1} = f(x_k,u_k)$ where $f: X\times U \rightarrow X$ and $X$ and $U$ are the state space and the control space, respectively.
We consider the motion planning problem from a start configuration, $x_{\mathrm{start}} \in X_{\mathrm{free}}$, to a configuration in the goal set, $X_{\mathrm{goal}} \subseteq X_{\mathrm{free}}$, where $X_{\mathrm{free}} \subseteq X$ is the set of valid robot states.
\begin{problem}[Motion planning problem]
Given $x_{\mathrm{start}} \in X_{\mathrm{free}}$ and $X_{\mathrm{goal}} \subseteq X_{\mathrm{free}}$, find a sequence of controls, $\bar{u} = [\bar{u}_0,\dots, \bar{u}_{N-1}]$, such that $x_N \in X_{\mathrm{goal}}$ under robot dynamics $x_{k+1} = f(x_k, u_k)$.
\end{problem}
We assume that the planning algorithm has access to (i) an approximate model of the system, $\hat{f}: X\times U \rightarrow X$, to simulate the outcome of a state-and-action pair and (ii) a model deviation estimate, $\mde(x,u) \approx || f(x,u) - \hat{f}(x,u) ||$.
Because $\hat{f} \neq f$, the planner may find solutions that do not reach the goal or collide with the environment when executed.
A possible solution to this issue is to search for valid solutions under both the true and the approximate dynamics by assuming or estimating the properties of the model uncertainty \cite{Glen:trusted-domain, planning-belief-space, Pavone:robust-rrt}.
Another approach consists of optimistically planning with the approximated dynamics and triggering replanning after each action or when the executed trajectory deviates too much from the planned one \cite{CMAX,CMAX++,Peters:monte-carlo-path-planning}.
In this work, we follow this second approach.
In particular, we assume a safety function is available to stop the robot when it deviates from the nominal trajectory.
Such a \texttt{replan\&execute} paradigm is exemplified in Alg. \ref{alg:replan-and-execute}.

The \texttt{replan\&execute} paradigm can be combined with the \mde\ to plan for trajectories that avoid high-error transitions \cite{Mitrano:science-robotics, Faroni_RAL2023}.
However, it can become inefficient when there is a mismatch between the information embedded in the \mde\ function and the real environment.
For example, consider a planner that minimizes the \mde\ function over the trajectory, as in \cite{Faroni_RAL2023}.
Suppose that the optimal solution collides with the environment during the execution. 
Replanning by minimizing the same \mde\ function would return a solution similar to the previous one, and we would not expect the robot to make progress toward the goal.
To overcome this issue we need to change the structure of the motion planning problem, e.g., by updating the cost function with the new information.

Our goal is to adapt the planning problem by embedding online information gathered from real observations so that future re-plannings will avoid transitions whose actual error is inconsistent with the \mde\ estimate.
We do so by adapting the motion planner cost function and sampling distribution so that it will discourage transitions with a high probability of being unreliable, i.e., those with large expected error or similar to previous unreliable transitions.

\begin{algorithm}[tpb]
\DontPrintSemicolon
\SetKwFunction{sampleTo}{sampleTo}
\KwIn{$\x_{\mathrm{start}}$, $X_{\mathrm{goal}}$, $\hat{f}$, $\delta_{\mathrm{safe}}$ }
$x = x_{\mathrm{start}}$, $\hat{x} = x_{\mathrm{start}}$\;
\While{$x \not\in X_{\mathrm{goal}}$}
{
    $\bar{u} = \texttt{plan}(x,X_{\mathrm{goal}})$\; \label{alg:replan-and-execute:plan}
    \For{$\bar{u}_i$ \KwInterval $\bar{u}_0, \dots, \bar{u}_{N-1}$ }
    {
        $x = \texttt{execute}(\bar{u}_i)$\;
        $\hat{x} = \hat{f}(\hat{x},\bar{u}_i)$\;
        \If{$|| x - \hat{x} || \geq \delta_{\mathrm{safe}}$}
        {
            \KwBreak\; \label{alg:replan-and-execute:safety-fcn}
        }      
        
    }
 }
\caption{The \texttt{replan\&execute} paradigm.}
\label{alg:replan-and-execute}
\end{algorithm}

\section{Method} \label{sec:method}

\begin{figure*}[tbp]
	\centering
    \begin{overpic}[width=0.9\textwidth]{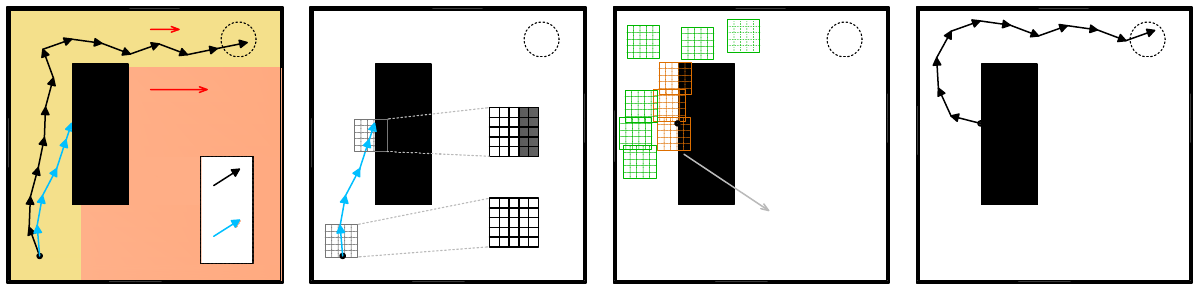} 
    \put (4,2.2) {\footnotesize$\x_{\mathrm{start}}$}
    \put (18,18) {\footnotesize$X_{\mathrm{goal}}$}
    \put (12.2, 22.2) {\tiny \RED{small drift}}
    \put (12.2, 17.2) {\tiny \RED{large drift}}
    \put (17,7.5) {\tiny planned}
    \put (16.9,2.9) {\tiny executed}
    \put (39,2.5) {\tiny {local env. of $\tau_{\tiny 1}^{\tiny E}$}}
    \put (39,10) {\tiny {local env. of $\tau_{\tiny 5}^{\tiny E}$}}
    \put (64.5,12) {\tiny transitions with} 
    \put (64.5,10) {\tiny context similar}
    \put (64.5,8) {\tiny to $\tau_5^E$ are deemed}
    \put (64.5,6) {\tiny as unreliable}
    \put (88.5,12) {\tiny new solutions tend} 
    \put (88.5,10) {\tiny to avoid unreliable}
    \put (88.5,8) {\tiny transitions according}
    \put (88.5,6) {\tiny to online data}
    \end{overpic}
	\caption{Sketch of the proposed method. From left to right: (i) the robot plans and executes a trajectory minimizing the expected error. It stops when it deviates too much from the nominal path.  (ii) every time it stops, we observe the actual error and the context of the executed transitions.
    (iii) we update the cost function and the sampling bias to avoid transitions similar to the unreliable ones.
    (iV) the new solution will likely avoid such transitions.} 
	\label{fig: example}
 \vspace{-0.5cm}
\end{figure*}
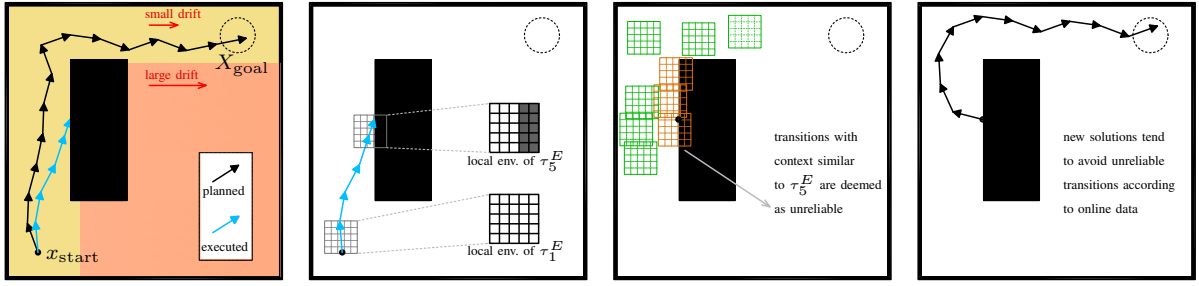

We can summarize the proposed method as follows:
\begin{enumerate}
    \item The motion planner searches for a path that minimizes the \mde\ function;
    \item The robot executes the trajectory; it stops if the deviation from the nominal path is larger than a safety threshold.
    \item At each safety stop, we measure the mismatch between the execution error of each transition and its \mde\ value. 
    We also compute the context of each executed transition, i.e., a representation of the environment in the proximity of the transition.
    \item We update the cost function to be used for the next replanning. 
    To do so, we classify previous transitions as unreliable by running an anomaly detection algorithm on the execution error.
    Then, we add a penalty term to transitions that are similar to previous unreliable transitions. The similarity function is based on the similarity between the local environments of the new and the executed transitions.
    This allows the planner to infer the probability that a transition is unreliable (i.e., anomalous) even when it is far from previous ones in the state-and-action space.
    This behavior is also enforced during sampling by leveraging an online learning bias strategy \cite{Faroni_RAL2023} that discourages sampling transitions with a high probability of being unreliable.
\end{enumerate}
For a better understanding of the approach, consider the simplified problem in Fig. \ref{fig: example}, where a point robot moves in a 2D environment and is subject to a drift to the right.
The magnitude of the drift depends on the robot position, as shown in the left image.
Our approach initially plans and executes a trajectory found by minimizing the nominal error estimate.
For this reason, the path lies in the low-drift (yellow) region.
During the execution, the robot deviates from the nominal trajectory and may collide with the obstacle, causing a halt.
Before replanning, we evaluate the local context of each executed transition (second image from the left).
The context contains information on the obstacle presence around the transition.
During the replanning, we use this information to infer the reliability of new simulated transitions.
For example, in the third image, the planner undersamples and assigns a large cost to transitions with a context similar to that of unreliable transitions (orange grids).
The result is that the new solution will be less likely to contain transitions with unfavorable context (right image).

The following sections illustrate the unreliability detection, the adaptive cost function, the context-based similarity, and the adaptive biased planner in detail.

\subsection{Detection of unreliable executed transitions} \label{sec:anomaly-detection}

Consider the executed transitions, $\mathcal{T}^E = \{\tau_1^E,\dots, \tau_M^E\}$ and their execution error $e(\tau_j^E) = || f(\tau_j^E.x, \tau_j^E.u) - \hat{f}(\tau_j^E.x, \tau_j^E.u) ||$.
We consider a transition to be unreliable if the predicted  and the actual execution error have an anomalous mismatch.
Thus, we run an anomaly detection algorithm on $e(\tau_j^E) \, \forall j=1,...,M$, and assign an anomaly label, $l^E_j$, to each transition such that:
\begin{equation}
    l^E_j = l(\tau_j^E) = \begin{dcases*}
        1& if $e(\tau_j^E)$ is anomalous\\
        0 & otherwise
    \end{dcases*}
\end{equation}
Anomalous transitions are those for which the \mde\ did not predict the execution error reliably.
We will therefore try to avoid using similar transitions during the next replannings.

\subsection{Adaptive cost function} \label{sec:adaptive-cost-fcv}

At each replanning, we minimize the cost function
\begin{multline}
    c(\tau) = \mde(\tau) + P_{\mathrm{anomaly}}(\tau) \, C + (1-P_{\mathrm{anomaly}}(\tau) ) \, \hat{e}(\tau)
    \label{eq:cost-fcn}
\end{multline}
where $C>0$ is a penalty term to discourage anomalous transitions, $P_{\mathrm{anomaly}}(\tau)$ is an estimate of the probability that $\tau$ will be anomalous if executed, and $\hat{e}(\tau)$ is an estimate of error $e(\tau)$ if $\tau$ will be executed.
The \mde\ term considers the nominal estimate of the model error. 
The second and the third term come into play after the first execution: the second term penalizes transitions 
with high probability of being anomalous/unreliable; the third one corrects the \mde\ estimate according to the error observed for similar transitions.

To compute $P_{\mathrm{anomaly}}(\tau)$ and $\hat{e}(\tau)$ for new, simulated transitions, we define a similarity measure, $S(\tau_i,\tau_j^E) \in [0,1]$, between transitions and use it as a proxy of the probability that $l(\tau_i) = l(\tau_j^E)$.
Then, we denote by $\mathcal{T}^E_{\mathrm{srt}}$ the sequence composed of the elements of $\mathcal{T}^E$ sorted in descending order according to $S(\tau,\tau_j^E)$ and  denote by $S_{\mathrm{srt}} = [S(\tau,\tau_{\mathrm{srt},1}^E),..., S(\tau,\tau_{\mathrm{srt},M}^E)]$ the corresponding similarity vector.
Thus, 
\begin{multline}
    P_{\mathrm{anomaly}}(\tau) = S_{\mathrm{srt},1}\, l(\tau_{\mathrm{srt},1}^E)+ (1-S_{\mathrm{srt},1} )  S_{\mathrm{srt},2} l(\tau_{\mathrm{srt},2}^E) \\+ ... + \prod_{k=1}^{M-1} (1- S_{\mathrm{srt},k} ) S_{\mathrm{srt},M} l(\tau_{\mathrm{srt},M}^E)
\end{multline}
$P_{\mathrm{anomaly}}$ approximates the probability that $\tau$ is unreliable by weighting all executed transitions according to the their similarity with $\tau$. 

We leverage the similarity measure also to compute $\hat{e}(\tau)$ as the weighted average of the observed residual errors:
\begin{equation}
    \hat{e}(\tau) = \frac{\sum_{j=1}^{M} S(\tau,\tau_{j}^E) \, e(\tau_j^E) }{\sum_{j=1}^{M} S(\tau,\tau_{j}^E)}
\end{equation}

\subsection{Context-based transition similarity} \label{sec:similarity}

Cost function \eqref{eq:cost-fcn} relies on the definition of the similarity function $S(\tau_i,\tau_j)$.
Previous works on sampling-based motion planning trying to leverage similarity between transitions or configurations typically relied on state-and-action  Euclidean distance \cite{Faroni_RAL2023,tsiotras:Epanechnikov}.
However, this only allows the planner to infer the properties of new transitions when they are in the local neighborhood of previous ones.
To overcome the locality issue, we introduce the idea of context similarity, i.e., we deem two transitions as similar if they have similar local environment features.
\rev{The definition of a suitable context-similarity function is problem-dependent.}
In our examples, we use a local grid centered at the robot end-effector pose and compute a context vector $\zeta$ of size $F+1$ such that
\begin{equation}
\label{eq:context-vector}
    \zeta_i = \begin{dcases*}
        \mde(\tau) & if $i=F+1$\\
        1 & if $i\leq F$ and the $i$th point of the local grid \vspace{-0.2cm}\\
          & \,\,\,\, is in collision with the environment \vspace{-0.1cm}\\
        0 & otherwise
    \end{dcases*}
\end{equation}
\rev{The similarity measure is then} $S(\tau_i,\tau_j) = e^{\frac{1}{2}||\zeta(\tau_i) -  \zeta(\tau_j)||} \in [0,1]$.
\rev{How to automatically define $S(\tau_i,\tau_j)$ for a given problem is an open question that we leave for future work.}

\subsection{Adaptive context-aware sampling bias} \label{sec:mab-rrt}

We embed our planning strategy in a kinodynamic sampling-based planner based on \MABRRT\ \cite{Faroni_RAL2023}.
\MABRRT\ runs multiple instances of \RRT\ sequentially and, at the end of each run, clusters previous transitions according to a reward function.
Then, in the next run, it uses the clusters, $\mathcal{C}_i$, as arms of a Multi-Armed Bandit algorithm.
MAB iteratively chooses from which cluster to draw the next transition, trading off the exploitation of high-reward (i.e., low-cost) clusters and less-explored regions.
This results in a more efficient sampling of the state space and better path quality.

We use \MABRRT\ as a planner in line \ref{alg:replan-and-execute:plan} of Alg. \ref{alg:replan-and-execute} with reward function $r(\tau) = - c(\tau)$.
Then, we modify the clustering strategy to leverage the online information gathered from the executed transitions.
We use the similarity function $S(\tau_i,\tau_j)$ to compute the clusters; then we compute the initial reward of the $i$th arm of the MAB as 
$$
    \bar{r}_i = \frac{1}{|\mathcal{C}_i|} \sum_{\tau_j \in \mathcal{C}_i } -\mde(\tau_j).
$$
Finally, we adjust the initial reward of each arm based on the previously executed transitions.
To do so, consider the set of clusters $\mathcal{C} = \{\mathcal{C}_1,..., \mathcal{C}_H \}$, their initial rewards $\bar{r} = \{\bar{r}_1,..., \bar{r}_H\}$, and the set of executed transitions $\mathcal{T}^E$.
First, we predict whether $\tau_j^E$ belongs to the $i$th cluster.
If so, we adjust $\bar{r}_i$ according to the probability that $\mathcal{C}_i$ contains anomalous transitions, i.e.:
\begin{equation}
\label{eq:poison-clusters}
    \bar{r}_i = \bar{r}_i \bigg( 1 - \frac{| \{ \tau \in \mathcal{T}^E \text{ s.t. } b_i(\tau)=1 \wedge l(\tau) =1 \} |}{ | \{ \tau \in \mathcal{T}^E \text{ s.t. } b_i(\tau)=1 \} |} \bigg)
\end{equation}
where $b_i(\tau) = 1$ if $\tau$ was predicted to belong to $\mathcal{C}_i$ and 0 otherwise, and $|\cdot|$ is the cardinality of a set.
The result of this operation is that sampling clusters associated with anomalous transition is discouraged during planning.

The combination of the biased sampling and the adaptive cost function \eqref{eq:cost-fcn} will avoid unreliable transitions during planning.
\rev{Our approach assumes the planner minimizes the cost function during each planning. Although \cite{Faroni_RAL2023} did not provide formal guarantees on asymptotic optimality, results in Sec. \ref{sec:results} suggest the cost decreases consistently with iterations during each replanning.}

\section{Results} \label{sec:results}

This section validates our approach in simulated 2D scenarios and a 7-DOF real-world manipulation problem, showing that it outperforms the baselines in terms of execution success rate and number of replannings to reach the goal.

\subsection{2D navigation example} \label{sec:2d-example}

\begin{figure}[tbp]
	\centering
    \hspace{-1.5cm}
    \begin{overpic}[trim = 0cm 0cm 0cm 0cm, clip, angle=0, width=0.78\columnwidth]{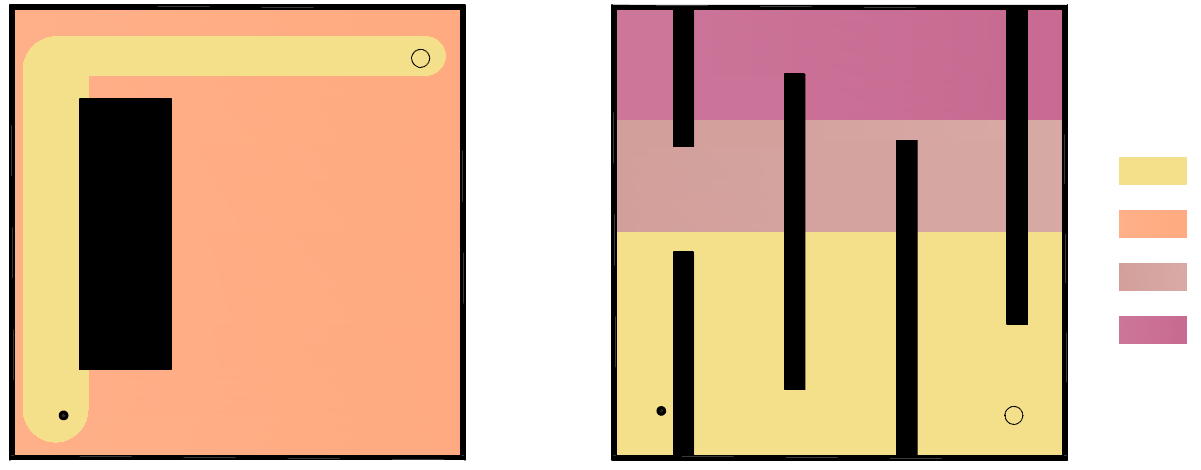} 
    \put (7,3) {\scriptsize$\x_{\mathrm{start}}$}
    \put (27,30) {\scriptsize$X_{\mathrm{goal}}$}
    \put (100,24) {\scriptsize $\delta = 0.006$}
    \put (100,19.25) {\scriptsize $\delta = 0.012$}
    \put (100,14.5) {\scriptsize $\delta = 0.018$}
    \put (100,9.75) {\scriptsize $\delta = 0.024$}
    \put (10,-5) {\footnotesize Scenario A}
    \put (60,-5) {\footnotesize Scenario B}
    \end{overpic}
    \vspace{0.3cm}
	\caption{Scenarios for the numerical analysis in Sec. \ref{sec:2d-example}.}
	\label{fig: 2d-cases}
  \vspace{-0.5cm}
\end{figure}
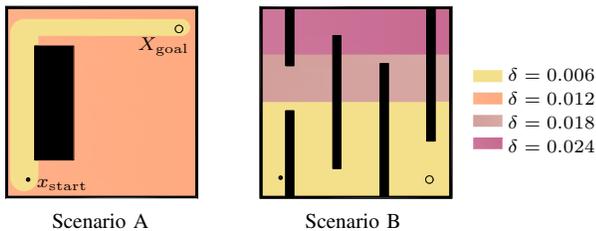

We consider the scenarios in Fig. \ref{fig: 2d-cases}.
The robot is a single integrator with a drift term $\delta>0$ such that $x_{k+1} = f(x_k,u_k) = x_k + T u_k + [\delta(x),0]^T$, $\X = [0,1]^2$, $\U\in[0.5,0.5]^2$, while the model used for planning is $x_{k+1} = \hat{f}(x_k,u_k) = x_k + T u_k$, where $T$ is the control duration.
The drift $\delta(x)$ depends on the robot position and is depicted in Fig. \ref{fig: 2d-cases}.
We consider $\mde = \delta(x)$. 

Note that there is a mismatch between the estimated error and the actual one because the robot stops as soon as it touches an obstacle. 
For example, in Scenario A, the low-error region near the left edge of the obstacle will actually result in a large execution error because the drift will drive the robot into contact with the obstacle.
The planner should learn to avoid such regions after a few executions, even though it encompasses the best nominal solution (i.e., it minimizes the \mde\ function).

To define the context variable, $\zeta$, we consider a 5x5 voxel grid centered in the robot frame.
For each new transition $\tau$, we set $\zeta_i = 1$ if the $i$th grid point is in collision with the environment and $\zeta_i = 0$ otherwise.
According to \eqref{eq:context-vector}, the last element of $\zeta$ is equal to $\mde(\tau)$.

We compare our context-aware algorithm, \CTXRRT, with its non-adaptive counterpart, \MABRRT\ minimizing the \mde\ function ($c(\tau) = \mde(\tau)$), and two variants for ablation:
\begin{itemize}
    \item \texttt{CTX-RRT-static-bias}, which adapts the cost function according to \eqref{eq:cost-fcn} but does not adapt clusters' rewards with the observed error;
    \item \texttt{CTX-RRT-static-cost}, which adapts the clusters' rewards according to \eqref{eq:poison-clusters} but not the cost function.
\end{itemize}

We evaluate the average number of replannings to reach the goal and the success rate (we report a failure if the robot does not reach the goal within 10 replannings).
Table \ref{tab: 2D-results} shows the results for 30 trials.
For both scenarios, \ctxrrt\ has the largest success rate and the smallest average number of replannings to reach the goal.
As expected, \mabrrt\ performs worse than all the planners because it tends to find similar paths near the obstacles even if such movements keep leading to a halt in execution.
The two variants, \texttt{CTX-RRT-static-cost} and \texttt{CTX-RRT-static-bias}, slightly outperform \mabrrt. However, they perform significantly worse than \ctxrrt, highlighting the importance of the combined adaptive bias and cost.
Results are consistent across the two scenarios, with an expected performance decay of all methods in Scenario B because of the higher complexity of the problem.

Note that such a combined adaptation is in line with the working principle of \mabrrt, which iteratively improves the solution by running multiple instances of \rrt\ and choosing the best solution. 
However, if the cost function does not change, the planner will use the nominal \mde\ to select the trajectory for execution.
On the other hand, if only the cost function changes, \mabrrt's performances decay because the MAB-based sampling is no longer effective with respect to the new cost function.

\begin{table}[tpb]
    \caption{Simulation results for the 2D scenarios.
	}	
	\label{tab: 2D-results}
	\centering
    \subfloat[][Scenario A]
	{
	$
	\begin{array}{lcc}
	\toprule
	& \text{Success rate} & \text{N. replannings} \\
	& \text{mean} & \text{mean (std.dev.)} \\
	\midrule
    \mabrrt & 0.40 & 5.9(1.9) \\
    \texttt{CTX-RRT-static-cost}  & 0.93 & 5.5(1.8) \\
    \texttt{CTX-RRT-static-bias} &  0.90  & 4.6(0.78) \\
    \ctxrrt &  \bf{0.97}  & \bf{3.5}(0.74) \\
	\bottomrule
	\end{array}
	$
 	\label{tab: 2D-results: a}
	}\\
     \subfloat[][Scenario B]
	{
	$
	\begin{array}{lcc}
	\toprule
	& \text{Success rate} & \text{N. replannings} \\
	& \text{mean} & \text{mean (std.dev.)} \\
	\midrule
    \mabrrt & 0.27  & 8.2(1.1) \\
    \texttt{CTX-RRT-static-cost}  & 0.54 & 7.9(0.55) \\
    \texttt{CTX-RRT-static-bias} &  0.27  & 8.0(1.2) \\
    \ctxrrt &  \bf{0.75}  & \bf{6.8}(0.98) \\
	\bottomrule
	\end{array}
	$
 	\label{tab: 2D-results: b}
	}
  \vspace{-1cm}
\end{table}

\subsection{Manipulation example} \label{sec:7D-example}

\begin{figure*}[tpb]
	\centering
    {\includegraphics[trim = 0cm 0cm 12cm 0cm, clip, angle=0, width=0.18\textwidth]{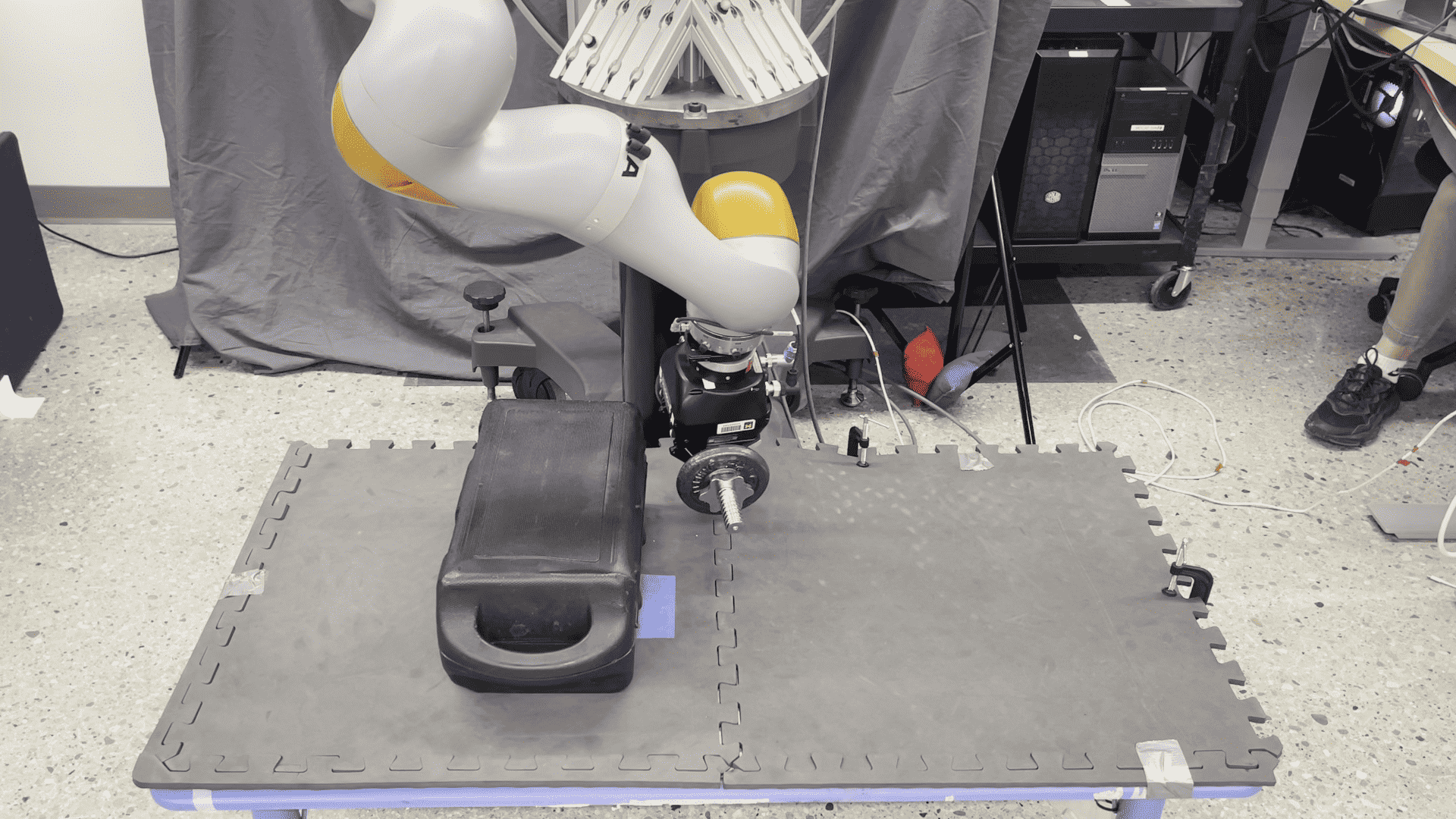}\,\,
    \includegraphics[trim = 0cm 0cm 12cm 0cm, clip, angle=0, width=0.18\textwidth]{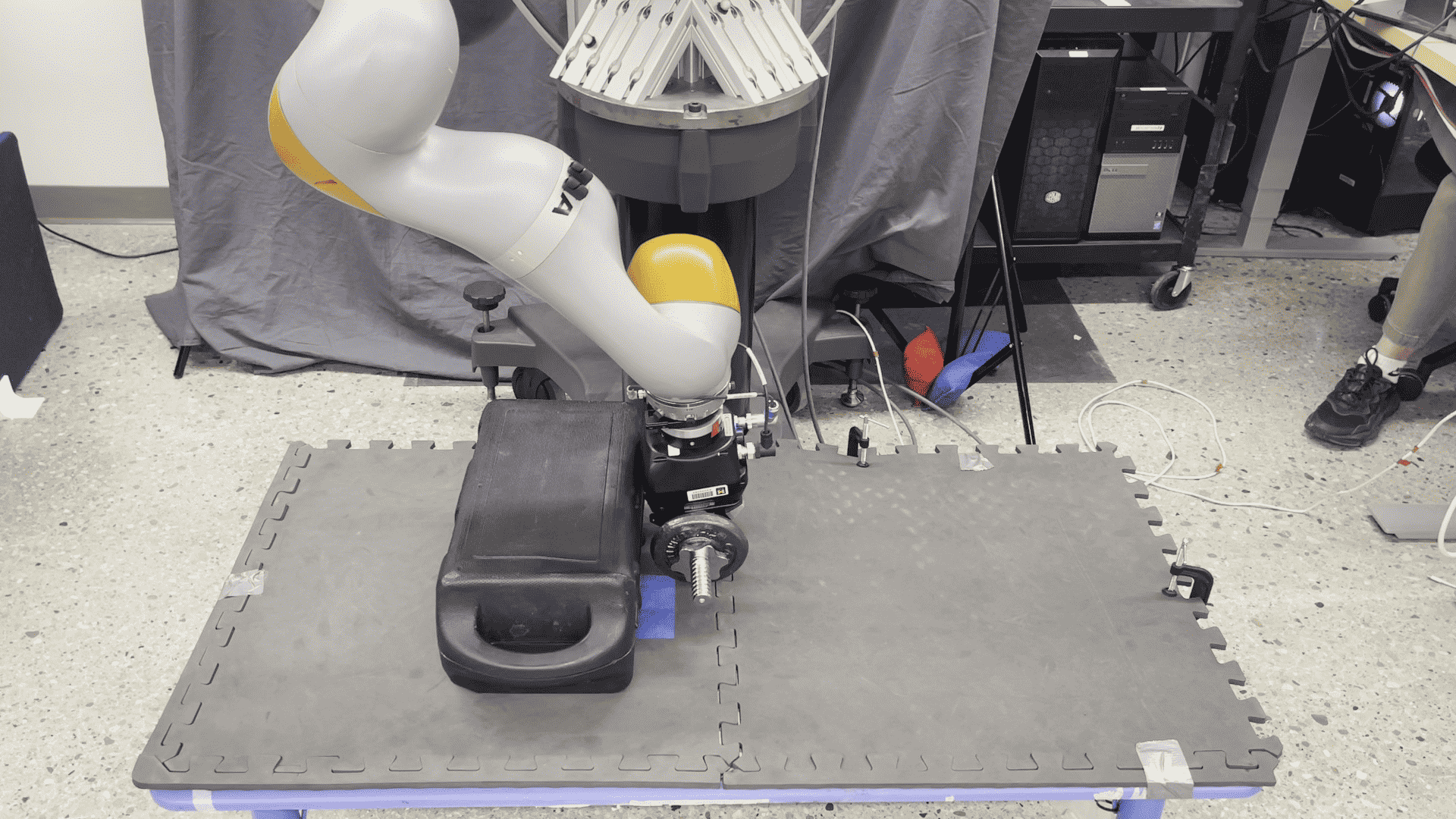}\,\,
    \includegraphics[trim = 0cm 0cm 12cm 0cm, clip, angle=0, width=0.18\textwidth]{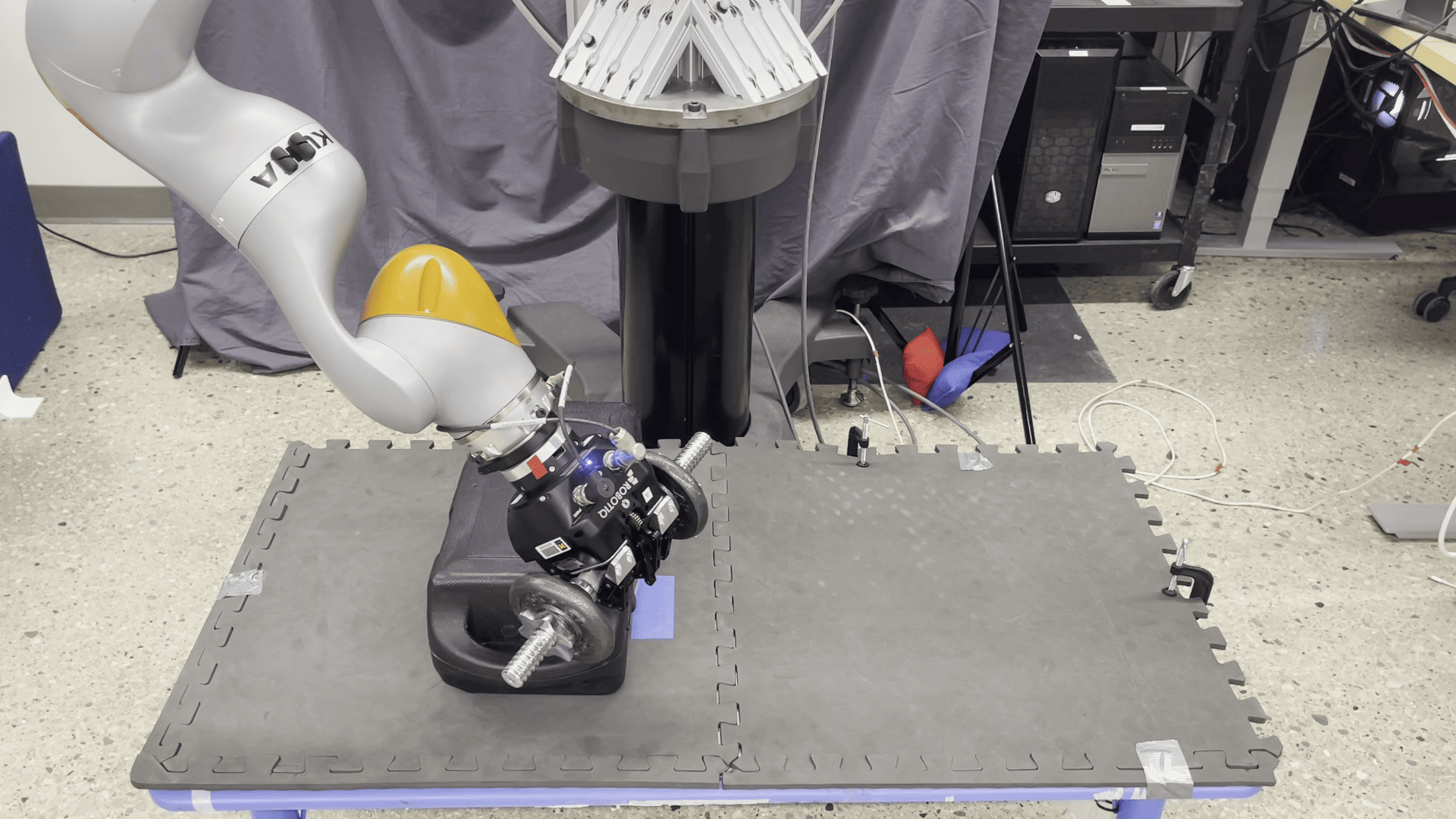}\,\,
    \includegraphics[trim = 0cm 0cm 12cm 0cm, clip, angle=0, width=0.18\textwidth]{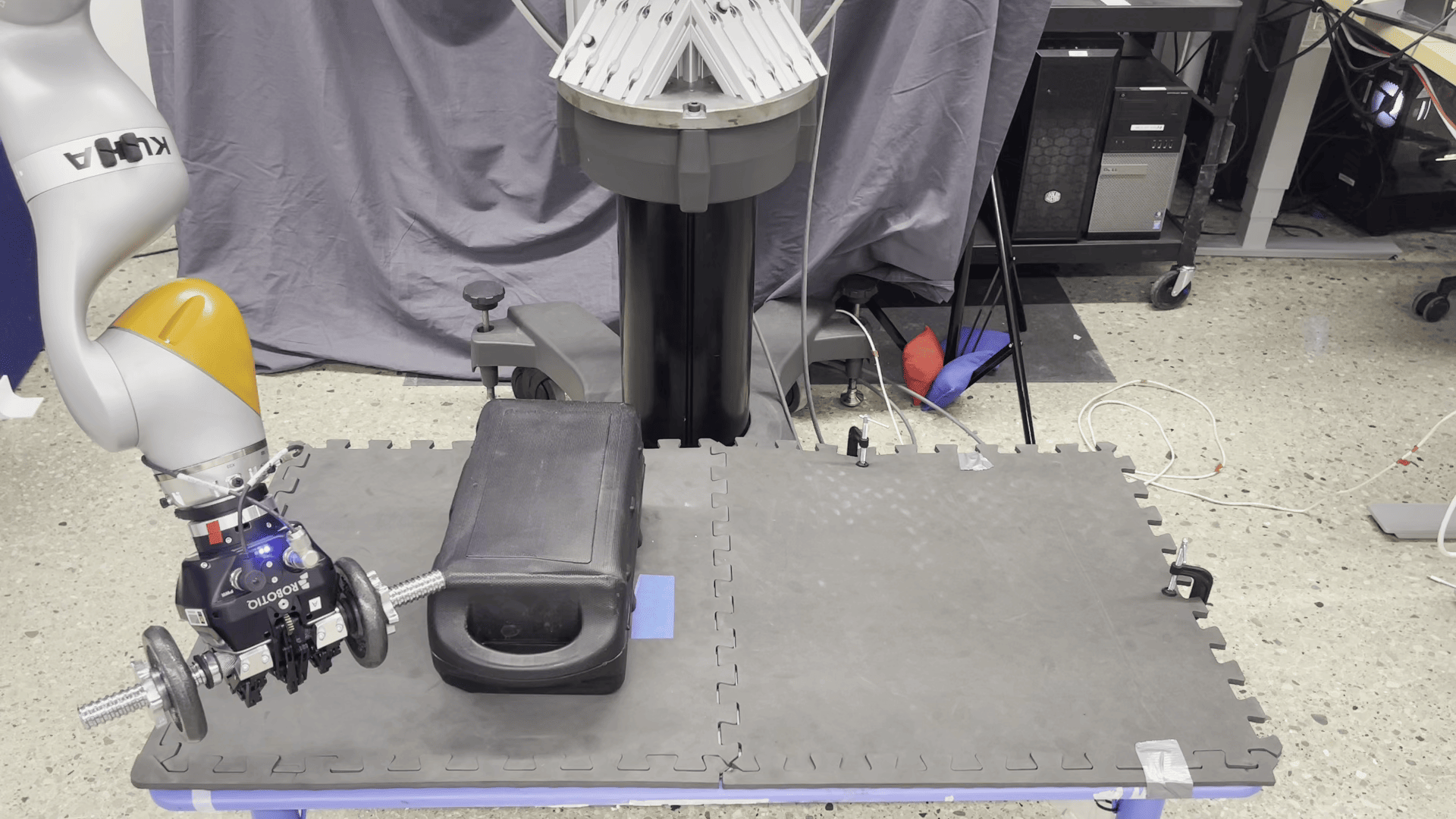}}\\
    \vspace{0.3cm}
    {\includegraphics[trim = 0cm 0cm 12cm 0cm, clip, angle=0, width=0.18\textwidth]{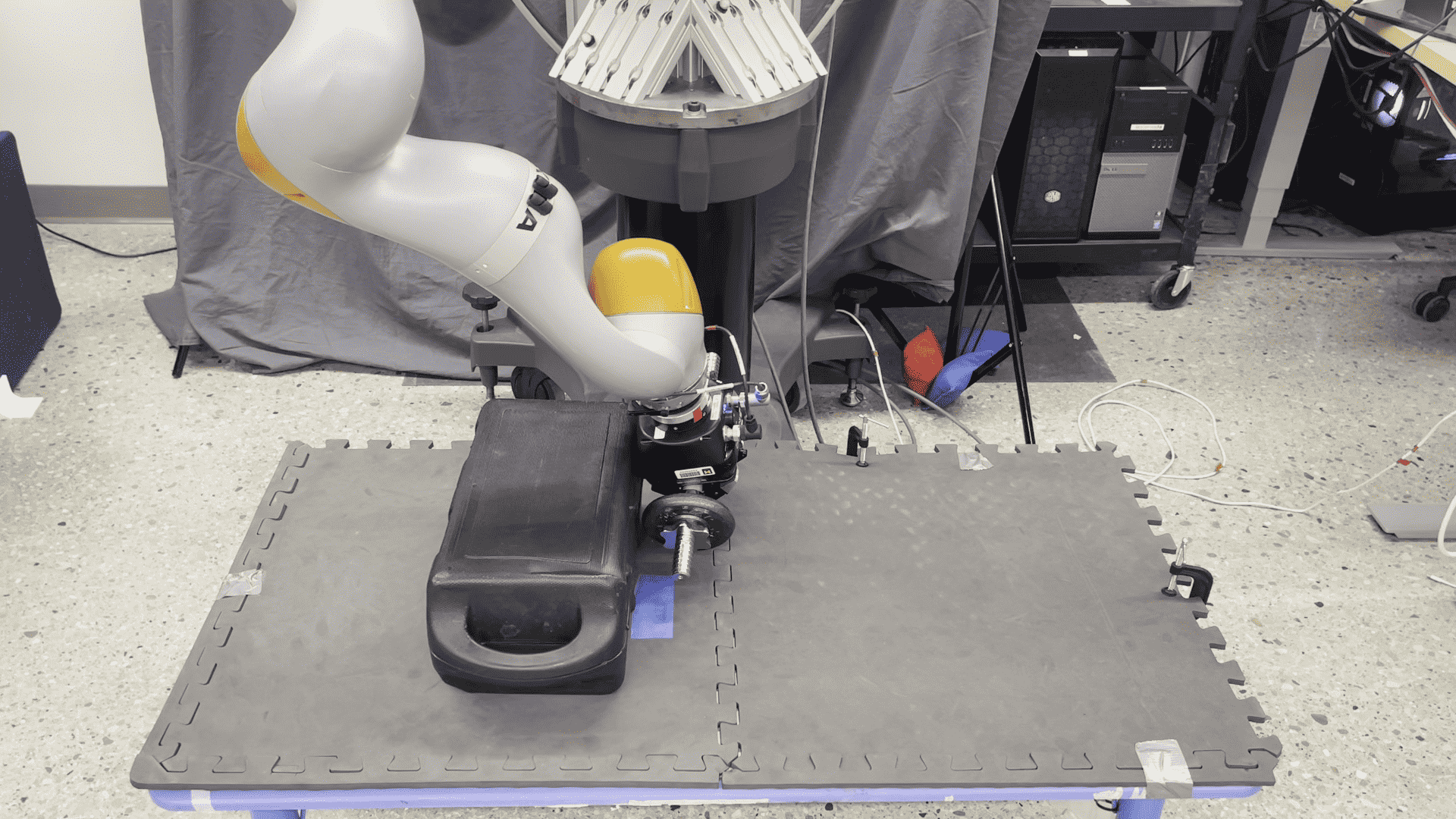}}\,\,
    \includegraphics[trim = 0cm 0cm 12cm 0cm, clip, angle=0, width=0.18\textwidth]{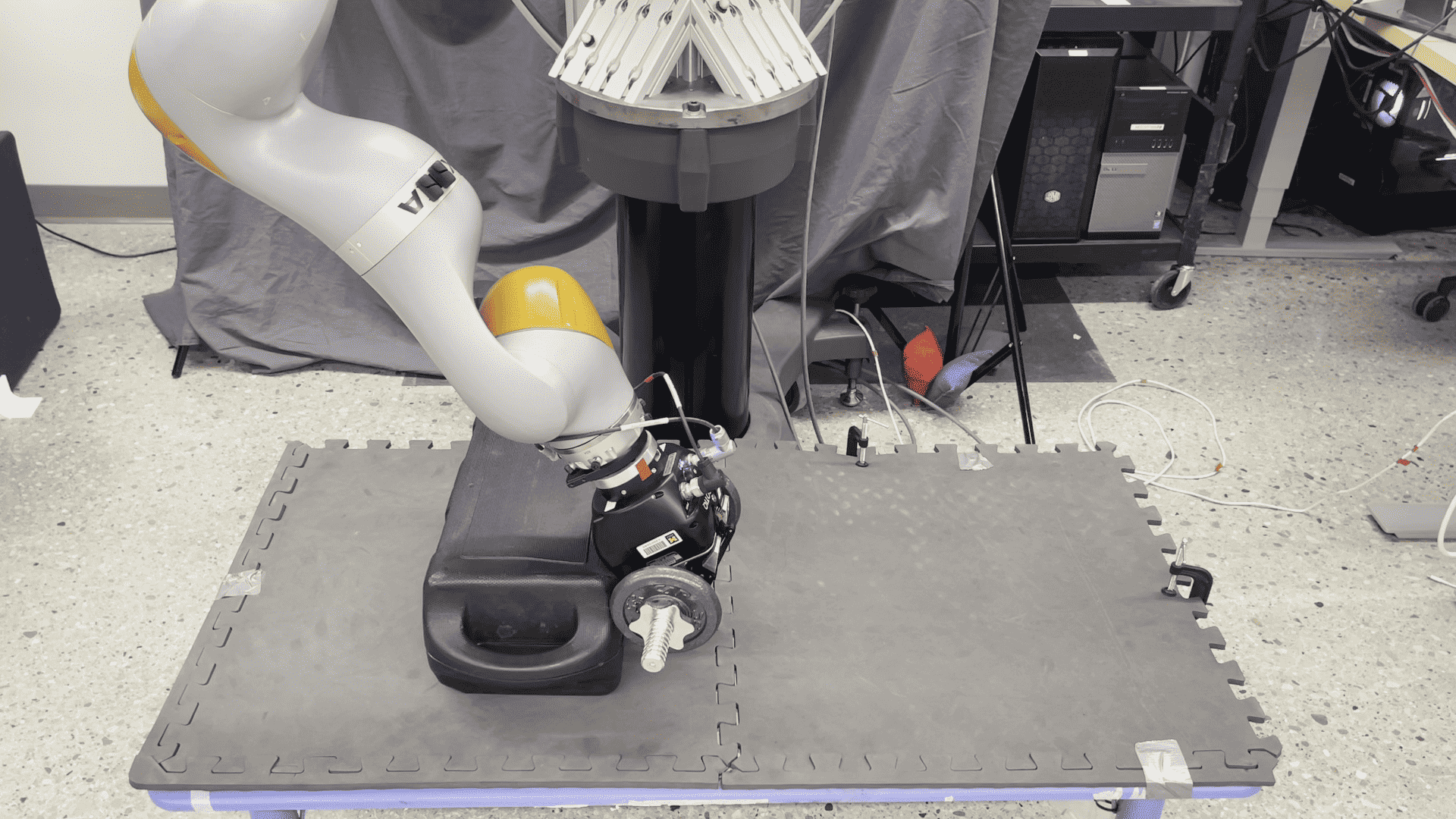}\,\,
    \includegraphics[trim = 0cm 0cm 12cm 0cm, clip, angle=0, width=0.18\textwidth]{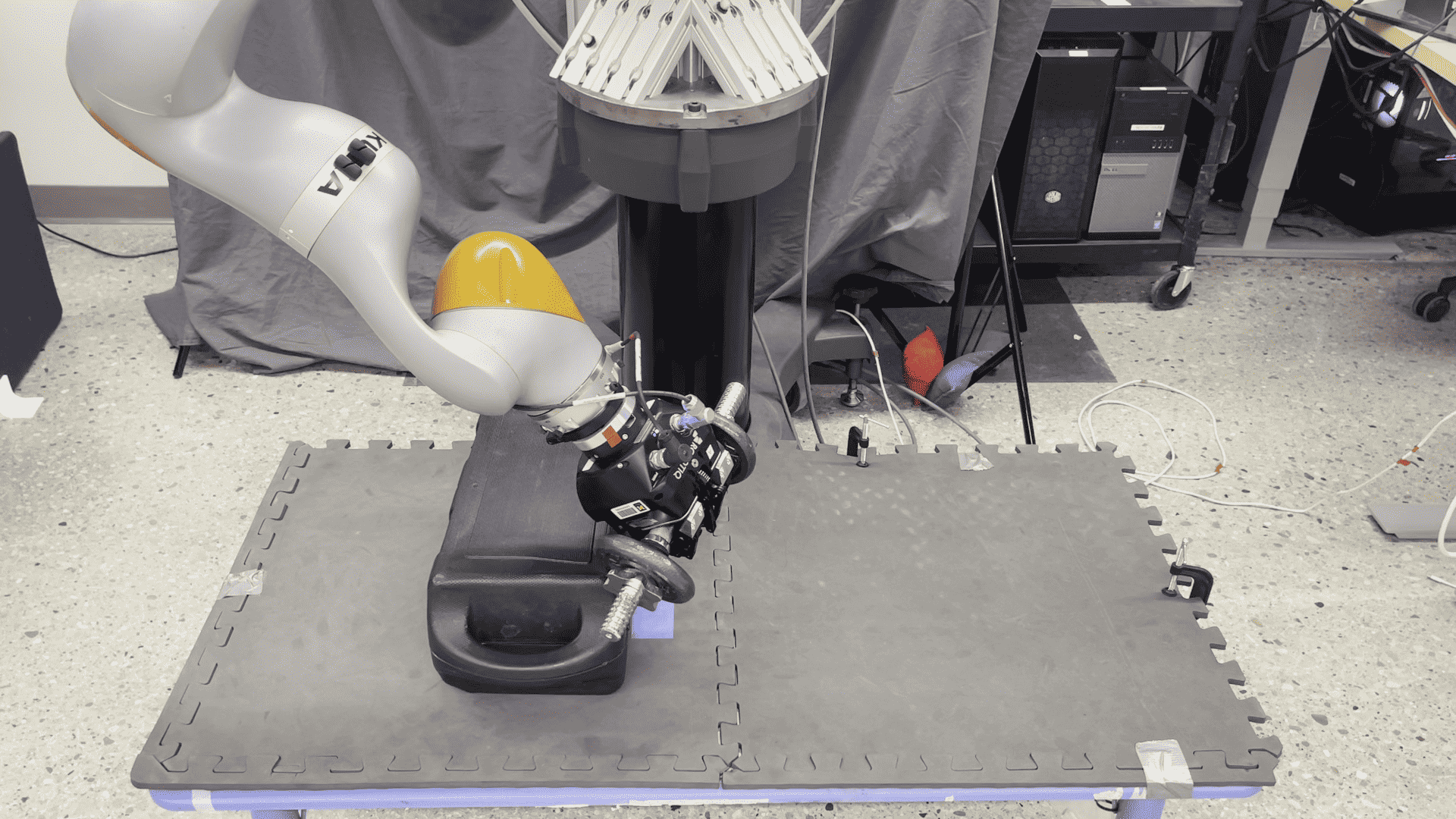}\,\,
    \includegraphics[trim = 0cm 0cm 12cm 0cm, clip, angle=0, width=0.18\textwidth]{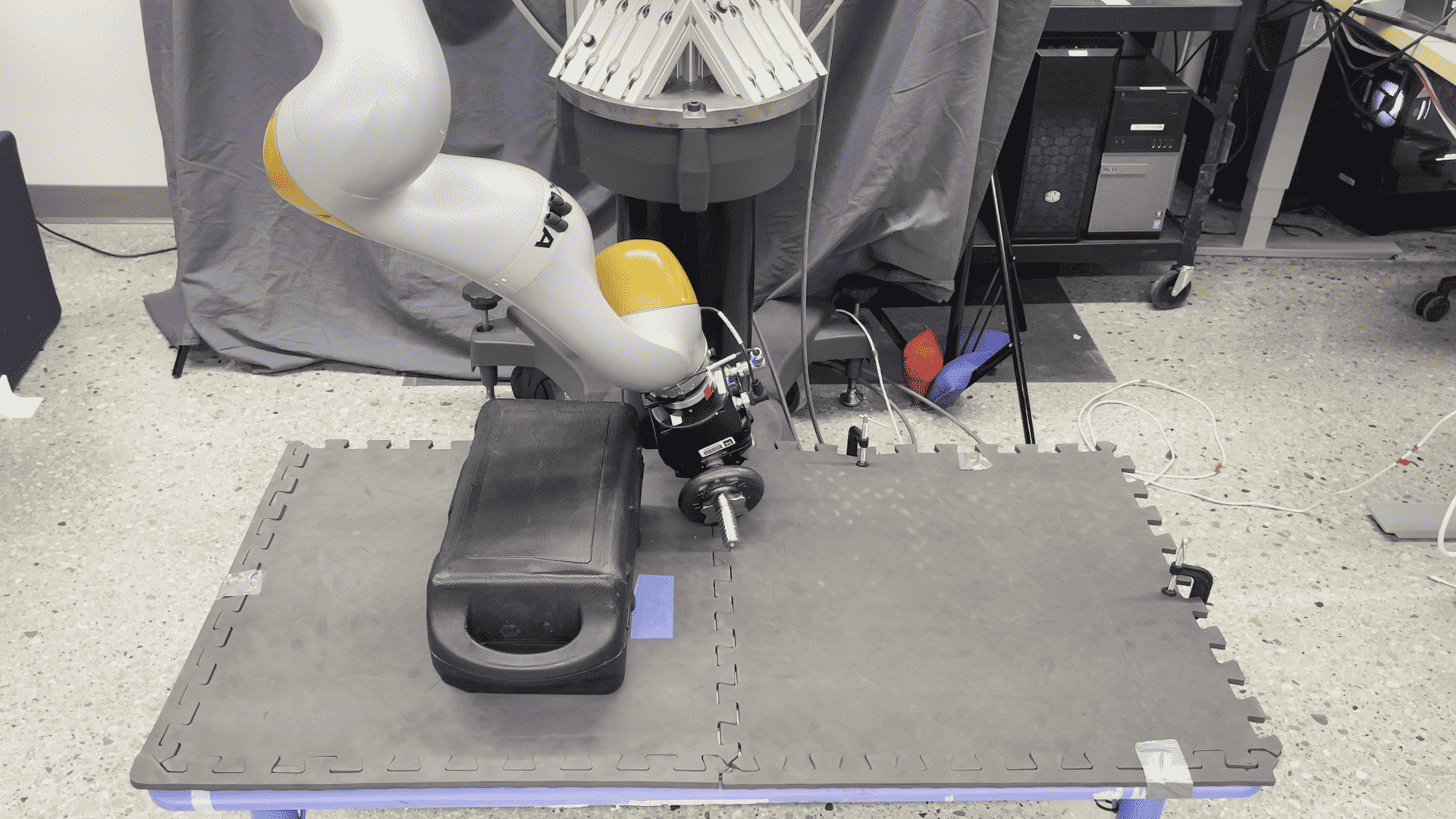}
	\caption{Examples of executions of two trajectories planned with \ctxrrt\ (top) and \mabrrt\ (bottom).}
	\label{fig: screenshots}
 \vspace{-0.5cm}
\end{figure*}

We demonstrate our approach on a manipulation task with uncertain dynamics.
We consider the tabletop application in Fig.
\ref{fig: victor-dumbbell}, where a 7-degree-of-freedom arm moves a dumbbell (3.6 kg) from a start to a goal pose on the table.
A joint-space impedance controller makes the robot compliant with the environment, yet it causes a significant trajectory-tracking error because of the large payload.
The state and control spaces are joint position and velocity, respectively.
The \mde\ function is an estimate of the robot Cartesian-space tracking error depending on the robot joint positions as described in Appendix A of \cite{Faroni_RAL2023}.
Note that both planners are aware of the obstacles (i.e., they perform collision checking on the simulated motions); yet collisions occur at runtime because of the trajectory tracking error discussed above.
Because of the tracking error, the robot may collide with the obstacles in the environment causing the robot to halt.
The \mde\ is able to estimate only the error owed to the robot dynamics.
Possible contacts with the obstacles and the inaccuracy of the \mde\ function cause a mismatch between the estimated and the measured Cartesian error.

To define the context variable, $\zeta$, we consider a rectangular cuboid centered on the robot tool center point and aligned with tool frame.
We consider a uniform grid of 25 points on each face of the cuboid.
Then, we set $\zeta_i = 1$ if at least half of the points on the $i$th face are in collision with the environment and $0$ otherwise.
Finally, the last element of $\zeta$ is equal to $\mde(\tau)$
according to \eqref{eq:context-vector}.

We evaluate the average number of replannings to reach the goal and the success rate (we report a failure if the robot does not reach the goal within 10 replannings).
Table \ref{tab: 7D-results} shows the comparison between \ctxrrt\ and \mabrrt\ (30 repetitions).
Note that the large robot compliance causes a low overall success rate with both planners.
Nonetheless, \ctxrrt\ significantly increases the success rate and reduces the average number of replannings needed to reach the goal.
In particular, \ctxrrt\ 
leverages the first executions to understand that motions too close to the obstacles lead to a large error even if it was not predicted by the \mde.
As a consequence, in the next replannings, it tends to avoid such motions and chooses paths that are less likely to result in a collision.
In comparison, \mabrrt\ keeps searching for the best solution according to the \mde's information, leading to multiple stops.
This is visible also in the accompanying video and in Fig. \ref{fig: screenshots}, which show executions with the two planners.
Initially, both planners find solutions that minimize the \mde\ function.
During the execution, the robot deviates from the nominal path and collides with the obstacle.
\ctxrrt\ uses the unexpected collisions to update the cost function and the sampling distribution and is eventually able to find a path in a different region of the space to avoid further collisions (top images).
In comparison, \mabrrt\ keeps finding similar solutions causing several halts (bottom images).

\begin{table}[tpb]
    \caption{Experimental results for the 7D scenario.
	}	
	\label{tab: 7D-results}
	\centering
	$
	\begin{array}{lcc}
	\toprule
	& \text{Success rate} & \text{N. replannings} \\
	& \text{mean} & \text{mean (std.dev.)} \\
	\midrule
    \mabrrt & 0.27 & 8.5(0.49) \\
    \ctxrrt & \bf{0.46}  & \bf{7.2}(0.82) \\
	\bottomrule
	\end{array}
	$
 \vspace{-0.5cm}
\end{table}

\section{Conclusions}\label{sec:conclusions}

We presented an adaptive planning strategy that can cope with model uncertainty and mismatches between the estimated and actual execution errors.
The approach adapts the cost function and sampling bias of a kinodynamic motion planner to discourage unreliable robot motions. 
Results on 2D examples and a 7D manipulation scenario showed that the approach improves the success rate during execution.
The current work relies on the definition of a context vector that captures the local environment features of each transition.
Future works will aim to learn an effective representation and similarity function for the transition context automatically.


\appendix

\subsection{Parameter tuning} \label{appendix:tuning}

This appendix describes the tuning of the parameters used to perform the experiments in Sec. \ref{sec:2d-example} and Sec. \ref{sec:7D-example}.

In all experiments, we tune the parameters needed to run \mabrrt\ as described in Appendix A.1 of \cite{Faroni_RAL2023}.
The anomaly detection algorithm (see Sec. \ref{sec:anomaly-detection}) uses a z-score anomaly detection with z=0.95 and we set $C=10$ in \eqref{eq:cost-fcn}.
In the 2D problems (Sec. \ref{sec:2d-example}), the context variable consists of a 5x5 grid with a uniform resolution equal to 0.05 and $\delta_{\mathrm{safe}}=0.05$.
In the 7D problem (Sec. \ref{sec:7D-example}), the context variable is a binary vector whose elements are associated with each face of rectangular cuboid of size 0.4m$\times$0.3m$\times$0.3m aligned with the x, y, and z axis of the tool frame and $\delta_{\mathrm{safe}}=0.2$ rad.




\newpage

\bibliographystyle{IEEEtran}
\bibliography{bib,bib_new}

\end{document}